\documentclass[10pt,twocolumn,letterpaper]{article}

\usepackage{iccv}
\usepackage{times}
\usepackage{epsfig}
\usepackage{graphicx}
\usepackage{epstopdf}
\usepackage{amsmath}
\usepackage{amssymb}
\usepackage[ruled,vlined]{algorithm2e}

\DeclareMathOperator*{\limt}{lim}
\newtheorem{definition*}{Definition}[section]
\newtheorem{theorem*}{Theorem}[section]

\usepackage[pagebackref=true,breaklinks=true,letterpaper=true,colorlinks,bookmarks=false]{hyperref}

\iccvfinalcopy 

\def\iccvPaperID{218} 

\ificcvfinal\fi
\begin{document}

\title{Zero-order Reverse Filtering}

\author{Xin Tao\footnotemark[1] \quad Chao Zhou\footnotemark[1] \quad Xiaoyong Shen\footnotemark[1] \quad Jue Wang\footnotemark[2] \quad Jiaya Jia\footnotemark[1]\\
\footnotemark[1]~~The Chinese University of Hong Kong\\
\footnotemark[2]~~Megvii Inc.
}

\maketitle

\begin{abstract}
In this paper, we study an unconventional but practically meaningful reversibility problem
of commonly used image filters. We broadly define filters as operations to smooth images
or to produce layers via global or local algorithms. And we raise the intriguingly problem if
they are reservable to the status before filtering. To answer it, we present a novel strategy to
understand general filter via contraction mappings on a metric space. A very simple yet
effective zero-order algorithm is proposed. It is able to practically reverse most
filters with low computational cost. We present quite a few experiments in the paper and
supplementary file to thoroughly verify its performance. This method can also be generalized
to solve other inverse problems and enables new applications.
\end{abstract}

\let\thefootnote\relax\footnotetext{Code will be available upon acceptance: \href{https://github.com/jiangsutx/DeFilter}{link}}
\section{Introduction}\label{sec:intro}
Image filtering is a fundamental building block of modern image processing and computer
vision systems. Recent advances in this field have led to new models to separate image
structure into different layers \cite{Karacan2013structure,Zhang2014rgf} or to remove
unwanted image structure
\cite{Gastal2011domain,Gastal2012adaptive,He2010guided,Xu2011l0,Xu2012ts,crossfield2013}
to satisfy the need of various tasks. The success provides the community deep
understanding of the capability of image filter.

In this paper, we broadly define ``filter'' as the operation, in either global
optimization or local aggregation way, to smooth images considering edge preserving
\cite{Tomasi1998bilateral,Farbman2008wls,Gastal2011domain,Gastal2012adaptive,He2010guided,Karacan2013structure,Perona1990scale},
texture removal \cite{Xu2012ts,Zhang2014rgf}, or other properties
\cite{Buades2005non,dabov2007bm3d,Rudin1992nonlinear,crossfield2013}. Unlike other work,
we aim to tackle an unconventional problem of
\begin{itemize}\vspace{-0.1in}
  \item removing part of or all filtering effect\vspace{-0.1in}
  \item without needing to know the exact filter in prior.\vspace{-0.1in}
\end{itemize}
We call our method Reverse Filtering or simply {\it DeFilter}.

The most related work to {\it DeFilter} is probably deconvolution
\cite{schuler2013machine,xu2010two}. But this line of research is by nature different --
deconvolution is to remove local linear convolution effect, while our goal is to study
the reversibility for even nonlinear/optimization based process. Another major
discrepancy is that we {\it do not} need to know the exact filter form in our method.
Therefore, our solver is not related to that of deconvolution at all.

\vspace{-0.1in}
\paragraph{Unique Property} Addressing this {\it DeFilter}
problem not only is intellectually interesting, but also enables many practical
applications. Since our method treats the filtering process as a black box, we can use it
to recover images processed by unknown but accessible operations.

\begin{figure}[t]
    \centering
        \includegraphics[width=1.0\linewidth]{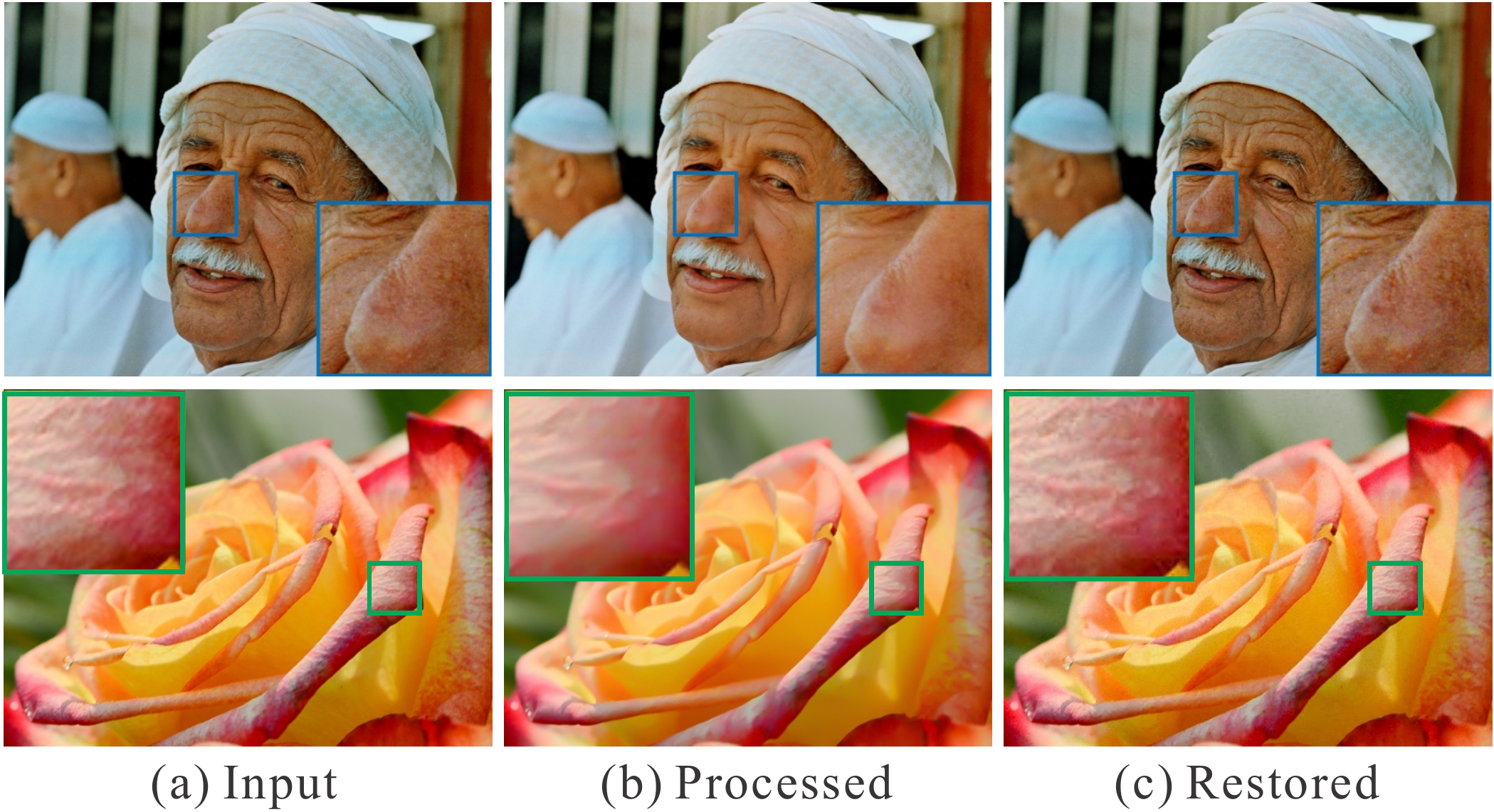}
\caption{(a) Original images. (b) Processed images by unknown retouching and
    denoising algorithms. (c) Our restored images.}\label{fig:app_retouch}
\end{figure}

Two examples are shown in Fig.~\ref{fig:app_retouch}. In the first row, the human face is
retouched by the Microsoft Selfie App using its ``Natural'' mode. We have no idea how it
is realized. In the second row, noise removal (operation ``denoise") in Photoshop
Express is applied. Again, we do not know the algorithm and its implementation details.
Interestingly, our method is still able to produce two results that are very close to the
original input before processed by the software, as shown in
Fig.~\ref{fig:app_retouch}(c). The lost patterns, \ie, freckles and wrinkles on the face
and texture of flower, are mostly recovered. To the best of our knowledge, this ability was
not explored or exhibited before in filter community.

\vspace{-0.1in}
\paragraph{Our Approach and Contribution}
Our new strategy is to understand the filtering process in metric space where filters are
considered as mappings. We provide detailed analysis and propose dividing the image
metric space into two sub ones $\Omega$ and $\bar{\Omega}$. In $\Omega$, a filtering
operation is strictly a {\it contraction mapping} (CM), which leads to a very simple {\it
zero-order} reverse filtering method to accurately remove unknown effect of filter with
low computation cost. Its theoretical correctness and convergence are guaranteed. For
filter in $\bar{\Omega}$ that is theoretically not invertible, our algorithm is still
effective to reduce its effect in practice, given the fact that most image filters are
designed to keep main structure and energy.

Our extensive study shows that the new {\it DeFilter} approach works very well on many
commonly employed local and global filters, including Gaussian filter, bilateral filter
\cite{Tomasi1998bilateral}, guided image filter \cite{He2010guided}, adaptive manifold
filter \cite{Gastal2012adaptive}, rolling guidance filter \cite{Zhang2014rgf}, BM3D
denoising \cite{dabov2007bm3d}, and many unknown filters incorporated in commercial
software. It also benefits non-blind deconvolution and super-resolution. We will present
a large amount of experimental results in the paper and supplementary material.



\section{Related Work}
\paragraph{Image Filter}\label{sec:related_filters}
Filtering is a basic procedure in computer vision and computational photography. Various
filters have been developed for many purposes, such as removing periodical/repetitive
textures \cite{subr2009edge,Xu2012ts}, reducing image noise
\cite{Buades2005non,dabov2007bm3d,crossfield2013}, or scale-aware/edge-aware smoothing
\cite{Tomasi1998bilateral,Gastal2011domain,Gastal2012adaptive,He2010guided,Perona1990scale,Zhang2014rgf},
to name a few.

Based on the supporting range used, filters can be categorized into local and global
ones. For local filters, a pixel value in the output image only depends on its close
neighbors. Representative methods include bilateral filter \cite{Tomasi1998bilateral},
guided filter \cite{He2010guided}, and anisotropic diffusion \cite{Perona1990scale}.
Global methods optimize energy functions defined over all or many pixels. Common
strategies include total variation (TV) \cite{Rudin1992nonlinear} and weighted least
squares \cite{Farbman2008wls}.

Depending on the property of continuity, most filters are continuous with respect to the
input image. Exception includes median-based filters \cite{huang1979fast,Zhang2014wmf}
and nearest-neighbor-based methods, such as BM3D \cite{dabov2007bm3d}.

\vspace{-0.1in}
\paragraph{Inverse Problems in Vision}
{\it DeFilter} belongs to the broad definition of {\it inverse problems} in computer
vision, where latent causal factors are estimated from observations. Typical inverse
problems include image denoising, deblurring, dehazing and super-resolution. For these
problems, prior knowledge is often used to regularize the solution space. For example,
the dark channel prior \cite{he2011single} is effective for image dehazing. Heavy-tail
image gradient prior \cite{RothB05foe,weiss2007makes}, spectral prior
\cite{goldstein2012blur} and color prior \cite{joshi2009image} are used for deblurring.
Learning-based approaches exploit information from external data. They include
sparse-coding-based methods for super-resolution \cite{yang2010image} and denoising
\cite{elad2006image}, as well as DNN-based super-resolution \cite{dong2014learning},
denoising \cite{schuler2013machine}, and deconvolution \cite{xu2014deep}.

We note these strategies do not fit our new task of {\it DeFilter}. Different image
filters have their respective properties, making it difficult to apply general image
priors or regularization. Learning-based methods need specific training for each image
filter involving parameters, which is also not considered in our solution. Unlike all
above methods, we resort to metric space mapping to tackle this new problem.

\begin{figure*}[ht]
\centering
    \includegraphics[width=1.0\linewidth]{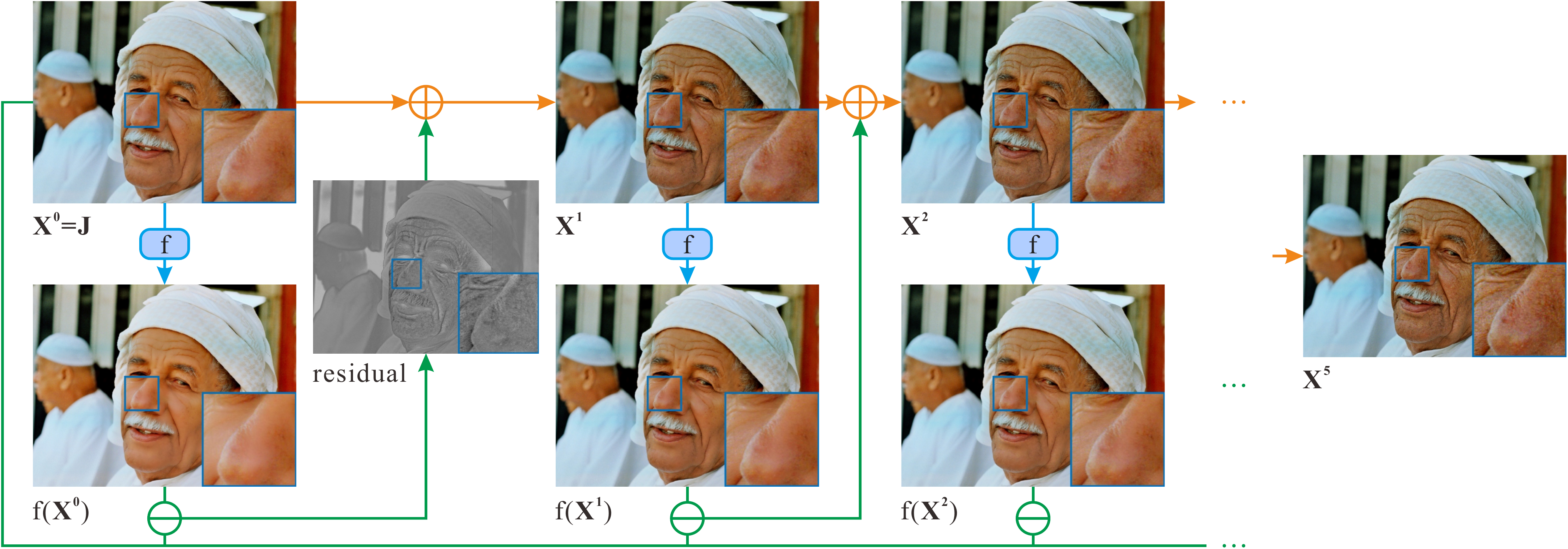}
    \caption{Illustration of our algorithm regarding operations in each iteration.
    }\label{fig:our_method}
\end{figure*}

\section{Reverse Filtering: Method and Conditions}
\label{sec:method} We first investigate the mathematical properties of general filters.
Without loss of generality, a filtering process can be described as
\begin{align}
  \mathbf{J} = f(\mathbf{I}),
\end{align}
where $\mathbf{I}$ and $\mathbf{J}$ are the input image and the filtering result. For
joint filtering methods \cite{He2010guided,crossfield2013}, we use $\mathbf{I}$ as the
guidance image so that $\mathbf{J}$ is still a function of $\mathbf{I}$. Our goal is to
estimate $\mathbf{I}$ without the need to compute $f^{-1}(\cdot)$.

\vspace{-0.1in}
\paragraph{Zero-order Reverse Filtering}
Our method is simple in terms of programming complexity. The {\it DeFilter} results can
be achieved using only a few lines of code. The main procedure is an iterative scheme,
which updates recovered images according to the filtering effect as
\begin{align}
    \mathbf{X}^{t+1}&=\mathbf{X}^{t}+\mathbf{J}^*-f(\mathbf{X}^{t}),\label{eq:defiler}
\end{align}
where $\mathbf{J}^*$ is the filtering result of image $\mathbf{I}^*$, with $\mathbf{J}^*
= f(\mathbf{I}^*)$. Both $\mathbf{I}^*$ and $f(\cdot)$ are unknown in this case.
$\mathbf{X}^{t}$ is the current estimate of $\mathbf{I}^*$ in the $t$-th iteration. It is
a {\it zero-order} algorithm because it does not require any derivatives.

To understand our algorithm, we use the illustration in Fig.~\ref{fig:our_method}. We
start from $\mathbf{X}^0=\mathbf{J}^*$, which is the filtered image, as initialization.
After re-applying the (unknown) filtering process to $\mathbf{X}^0$, details are further
suppressed. We then extract the residual $\mathbf{J}^*-f(\mathbf{X}^{0})$, which is the
difference involving a level of texture and details. The most unconventional step here
(highlighted by orange lines in Fig.~\ref{fig:our_method}) is that we add the residual
back to current estimate $\mathbf{X}^0$ to make resulting $\mathbf{X}^1$ contain these
details. Then we enter another iteration with similar steps. Intriguingly, $\mathbf{X}^t$
with increasing $t$ gets closer and closer to $\mathbf{I}^*$. For this example, only 5
iterations yield a good DeFilter result.

Despite the simple form, the proposed algorithm does not arbitrarily add back details.
Contrarily, in the following we prove for many filters this process is mathematically sound.

\subsection{Why Does This Simple Process Work?}
The form of Eq.~(\ref{eq:defiler}) can be considered as a kind of fixed-point iteration.
To facilitate analysis, we construct an auxiliary function $g(\mathbf{I})$ as
\begin{equation}
    g(\mathbf{I})=\mathbf{I}+(\mathbf{J^*}-f(\mathbf{I})). \label{eq:func_g}
\end{equation}
Eq.~(\ref{eq:func_g}) is equivalent to Eq. (\ref{eq:defiler}) for
$\mathbf{X}^{t+1}=g(\mathbf{X}^{t})$. To understand it better, we need {\it contraction
mapping} defined below to find internal relationship.

\begin{definition*}[Contraction Mapping]
    On a metric space $(\mathbb{M},d)$, a mapping $T:\mathbb{M}\rightarrow\mathbb{M}$
    is a \emph{contraction mapping}, if there exists a constant $c\in[0,1)$ such that
    $d(T(x),T(y))\leq c\cdot d(x,y)$ for all $x,y\in\mathbb{M}$.
\end{definition*}

With this definition, we take the mapping $T(\cdot)$ as $g(\cdot)$ in
Eq.~(\ref{eq:func_g}). In our case, $\mathbb{M}=\mathbb{R}^{m\times n}$ is the image
space of $m\times n$ pixels and $d(x,y)=\|x-y\|$ is the Euclidean distance. It is easy to
verify that $(\mathbb{M},d)$ is a complete metric space. With this definition, our
algorithm can be explained by following {\it Banach Fixed Point Theorem}.

\begin{theorem*}[Banach Fixed Point Theorem]
    Let $(\mathbb{M},d)$ be a non-empty complete metric space with a contraction mapping $T:
    \mathbb{M}\rightarrow \mathbb{M}$. $T$ admits a unique fixed-point $x^*$ in
    $\mathbb{M}$ (i.e., $T(x^*) = x^*$).

    Further, $x^*$ can be found in the following way. Start with arbitrary state $x_0$
    in $\mathbb{M}$ and define a sequence $\{x_n\}$ as $x_n = T(x_{n-1})$. When it
    converges, $\limt_{n\rightarrow\infty} x_n=x^*$.
\end{theorem*}

With this theorem, our algorithm can be understood as actually constructing an image
sequence $\{\mathbf{X}^{t}\}$, defined as $\mathbf{X}^{t}=g(\mathbf{X}^{t-1})$. If it
satisfies the sufficient condition, it will converge to the unique value $\lim_{t\to
\infty} \mathbf{X}^t=\mathbf{X}^*$. Put differently, the initial $\mathbf{X}^0$ is
processed by Eq.~(\ref{eq:defiler}) iteratively and finally reaches
$f(\mathbf{X}^*)\approx \mathbf{J}^*$.

Note the sufficient condition that theorem holds is that $g(\mathbf{I})$ forms a {\it
contraction mapping}, expressed as
\begin{align}\label{eq:rev_cond}
    \|g(\mathbf{I}_a)-g(\mathbf{I}_b)\|
    &=\|[\mathbf{I}_a-f(\mathbf{I}_a)]-[\mathbf{I}_b-f(\mathbf{I}_b)]\| \nonumber\\
    &\leq t\cdot\|\mathbf{I}_a-\mathbf{I}_b\|, \quad t\in[0,1)
\end{align}
For linear filters, the condition further reduces to
\begin{align}\label{eq:rev_cond_linear}
    \|\mathbf{I}-f(\mathbf{I})\|&\leq t\cdot\|\mathbf{I}\|. \quad t\in[0,1)
\end{align}

We analyze the contraction mapping condition in the following regarding different forms
of filters. The conclusion is vastly beneficial to the community -- {\it several filters
satisfy this condition completely}. For others, even it does not hold, our {\it
zero-order reverse filtering} still works to a decent extend to produce satisfying
results empirically.

\subsection{Convolutional Filter}
\begin{figure*}[ht]
\centering
    \includegraphics[width=1.0\linewidth]{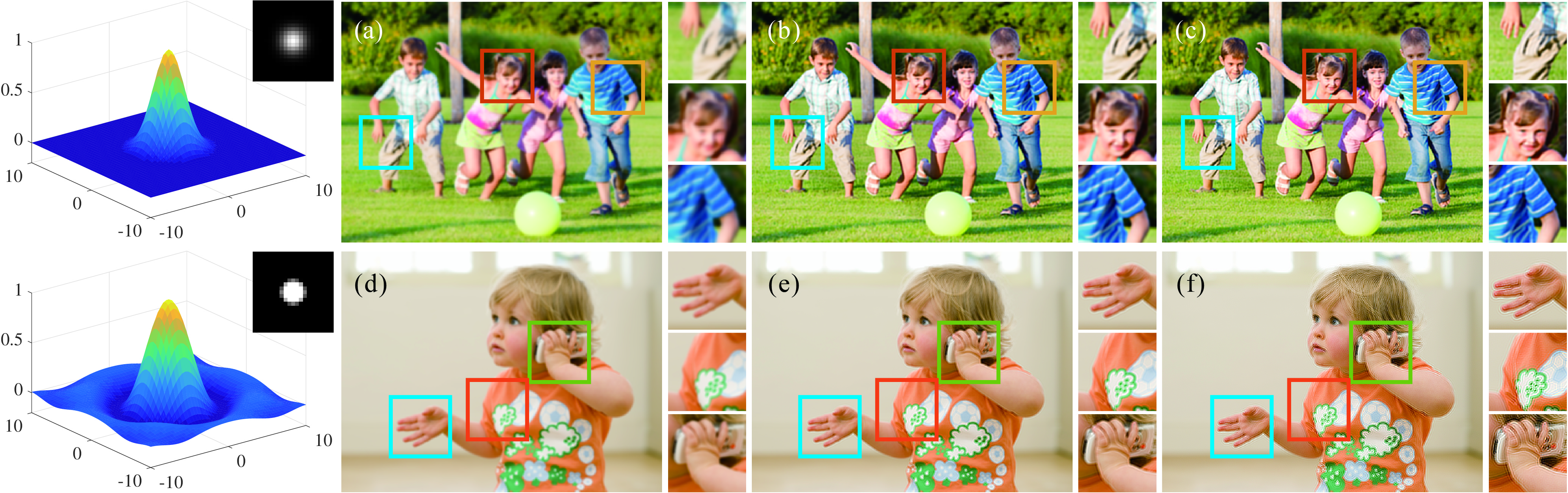}
    \caption{\textbf{Reversing convolutional filter.} Kernels and their
    power spectrum are shown in the first column.
    (a) Image blurred by a Gaussian kernel ($\sigma=2$, kernel size $21\times21$).
    (b)-(c) Results after 5 and 20 fixed-point iterations.
    (d) Image blurred by a disk kernel ($r=3$, kernel size $21\times21$).
    (e)-(f) Results after 5 and 15 iterations. Note that high-frequency
    artifacts start to appear in the 15th iteration.
    }\label{fig:subspace_iter}
\end{figure*}

We start from commonly employed convolutional filter expressed as
\begin{align}
  \mathbf{J} = f(\mathbf{I})=\mathbf{I}\ast \mathbf{K},
\end{align}
where $\ast$ is the convolution operator and $\mathbf{K}$ is the convolution kernel. For
ideal Gaussian filter (infinitely large support) and Tikhonov-regularized $L_2$-norm image smoothing filter, the spectrum
of the kernel contains only real positive numbers, i.e., $\hat{K}_p\in(0,1]$. Therefore,
\begin{align}\label{eqn:gauss}
  \|\mathbf{I}\ast(\mathbf{K}_\delta-\mathbf{K})\|=\|\mathbf{\hat{I}}\cdot(\mathbf{1}-\mathbf{\hat{K}})\|
  \leq t\cdot\|\mathbf{\hat{I}}\|=t\cdot\|\mathbf{I}\|
\end{align}
where $\hat{x}$ denotes Fourier transform, $\cdot$ is point-wised product, $\mathbf{1}$
is an all-one matrix and $p$ indexes pixels. The inequality holds when choosing
$t=1-\min_p \hat{K}_p<1$, which satisfies the condition in Eq.
(\ref{eq:rev_cond_linear}). It means these filters can be strictly reversed using
fixed-point iteration, if disregarding small numerical errors. As shown in
Fig.~\ref{fig:subspace_iter}(a)-(c), the DeFilter quality improves consistently in
iterations.

For other types of kernels, such as line and disk kernels
(Fig.~\ref{fig:subspace_iter}(d-f)), $\mathbf{\hat{K}}$ may contain zero, negative or
complex values that make $\|1-\hat{K}_p\|\geq 1$. In this case, Eq.
(\ref{eq:rev_cond_linear}) cannot satisfy arbitrary $\mathbf{I}$. But for
$\Omega=\{p\mid\|1-\hat{K}_p\|<1\}$, which is the set of frequency components that
satisfy Eq. (\ref{eq:rev_cond_linear}), the following inequality still holds.
\begin{align}\label{eq:freq_cond}
  \|\mathbf{\hat{I}}_\Omega\cdot(\mathbf{1}_\Omega-\mathbf{\hat{K}}_\Omega)\|\leq t\cdot\|\mathbf{\hat{I}}_\Omega\|,
\end{align}
where subscript $\Omega$ denotes that subset from the original image, which only contains
frequency points in $\Omega$. $t$ is set to $\max_p\|1-\hat{K}_p\|$ in this case, such
that $t<1$. Due to the linear property of convolution filter, fixed point iteration can
be split into two independent sequences as
\begin{align}\label{eq:two_subseq}
  \mathbf{X}^{t+1}=\mathbf{X}^{t+1}_\Omega+\mathbf{X}^{t+1}_{\bar{\Omega}}=g(\mathbf{X}^{t}_\Omega)+g(\mathbf{X}^{t}_{\bar{\Omega}}),
\end{align}
where $\bar{\Omega}$ denotes the complement of $\Omega$. According to previous analysis,
$\{\mathbf{X}^t_\Omega\}$ guarantees to converge to the unique solution
$\mathbf{X}^*_\Omega$, while $\{\mathbf{X}^t_{\bar{\Omega}}\}$ could oscillate or
diverge.

Fortunately, if we look at the spectrum of kernels, $\Omega$ region covers almost all
low-frequency components. For both Gaussian and disk kernels in
Fig.~\ref{fig:subspace_iter}, $\Omega$ region corresponds to the frequency whose power is
greater than 0, which is the majority. Meanwhile, for natural images, the low frequency
components dominate their energy. It means that {\em most of useful energy of the
original image can be recovered using fixed-point iterations}.

In the algorithm level, for the first a few iterations, reverse filtering adds back a lot
of details with $\{\mathbf{X}^t_\Omega\}$, which is the majority, dominating the process.
Excessive iterations (over 10) does not change $\{\mathbf{X}^t_\Omega\}$ much for its
near convergence and contrarily amplifies $\{\mathbf{X}^t_{\hat{\Omega}}\}$ with
high-frequency artifacts (Fig.~\ref{fig:subspace_iter}(d-f)). It is worth mentioning that
the divergent part contains filter-specific information.

\subsection{General Linear Filters}
Similar analysis holds for general linear filter of
\begin{align}
  \mathbf{J} = f(\mathbf{I})=\mathbf{A}I,
\end{align}
where $I$ is the vectorized image $\mathbf{I}$, and $\mathbf{A}$ is a square matrix
corresponding to linear filter. Singular value decomposition of $(\mathbb{I}-\mathbf{A})$
gives
\begin{align}
  \mathbb{I}-\mathbf{A} = \mathbf{U}\mathbf{S}\mathbf{V}^*,
\end{align}
where $\mathbb{I}$ is an identity matrix, superscript $*$ means conjugate transpose,
$\mathbf{U}$ and $\mathbf{V}$ are unitary matrices, and $\mathbf{S}$ is a diagonal matrix
containing singular values. We put these singular values with their squares less than 1
into set $\Omega$ where $\Omega=\{p\mid|diag(\mathbf{S})_p|^2<1\}$. We further project
vectorized image $X$ into two orthogonal subspaces as
\begin{align}\label{eq:subspace_linear}
  X_\Omega = \mathbf{V}\mathbf{D}_\Omega\mathbf{V}^*X, \quad  X_{\bar{\Omega}} =
  \mathbf{V}\mathbf{D}_{\bar{\Omega}}\mathbf{V}^*X,
\end{align}
where $\mathbf{D}_\Omega$ is a diagonal matrix, and $diag(\mathbf{D}_\Omega)_p$ is 1 if
$p\in\Omega$, otherwise it is set to 0.
$\mathbf{D}_{\bar{\Omega}}=\mathbb{I}-\mathbf{D}_\Omega$. We thus have
\begin{align}\label{eq:subspace_cond_linear}
  \|(\mathbb{I}-\mathbf{A})X_\Omega\|
  &=X^*\mathbf{V}\mathbf{D}^*_{\Omega}\mathbf{S}^*\mathbf{S}\mathbf{D}_{\Omega}\mathbf{V}^*X\nonumber\\
  &=X^*\mathbf{V}\mathbf{\Lambda}_{\Omega}\mathbf{V}^*X\leq t\cdot\|X\|,
\end{align}
where
$\mathbf{\Lambda}_{\Omega}=\mathbf{D}^*_{\Omega}\mathbf{S}^*\mathbf{S}\mathbf{D}_{\Omega}$
is a diagonal matrix containing only squared singular values in $\Omega$. It is easy to
verify that the inequality holds when choosing $t=\max_{p\in\Omega}
|diag(\mathbf{S})_p|^2<1$. Therefore, similar to Eq. (\ref{eq:two_subseq}), iterations
can be considered as sum of two independent subsequences in orthogonal subspaces, and
subsequent $\{\mathbf{X}^t_\Omega\}$ can strictly converge to optimal values. In fact,
the two separated regions in frequency domain (Fig.~\ref{fig:subspace_iter}) are special
forms of the orthogonal subspaces we derive in Eq. (\ref{eq:subspace_linear}).

The observation that $\Omega$ subspace contains the major energy of a natural image is
{\it not} by chance. Since most commonly used linear filters are designed to suppress or
remove unwanted components like noise and texture while retaining main structures,
original structure is mostly included in $\Omega$, whose reversibility is ensured by Eq.
(\ref{eq:subspace_cond_linear}). We will validate it extensively in experiments.

\subsection{Other Common Filters}
\begin{figure*}[ht]
\centering
    \includegraphics[width=1.0\linewidth]{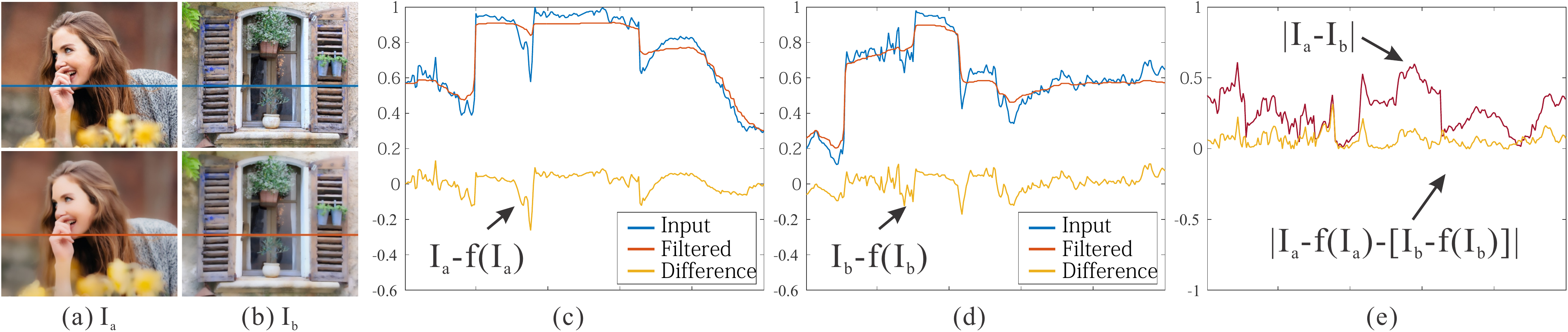}
    \caption{\textbf{Visualization of reversibility}. (a)-(b) Input and filtered images by WLS
    \cite{Farbman2008wls}, where the same scan-line pixels are selected as 1D signals.
    (c)-(d) 1D signals of input (\textbf{blue}), filtered (\textbf{red})
    and their difference (\textbf{yellow}). (d) Difference between noise components
    (\textbf{yellow}) and difference between input images (\textbf{red}).}\label{fig:contraction}
\end{figure*}

We extend the idea of considering filter as operations in two separate subspaces. For
most image smoothing filters, they are designed to remove the ``noise'' component. In
natural images, noise and texture are small-scale components compared to dominant
structure and edges. Taking weighted-least-square \cite{Farbman2008wls} as an example, we
apply it to two random images shown in Fig.~\ref{fig:contraction}. The ``noise''
components $\mathbf{I}_a-f(\mathbf{I}_a)$ and $\mathbf{I}_b-f(\mathbf{I}_b)$ are the
yellow curves in Fig.~\ref{fig:contraction}(c-d), respectively. In
Fig.~\ref{fig:contraction}(e), the absolute difference between the two noise components
(in yellow) is much smaller than the absolute difference between the two unfiltered
images (in red), which make the filter nearly satisfy the {\it contraction mapping}
condition in Eq.~(\ref{eq:rev_cond}) in practice. In the experiment section, we will
verify this on lots of filters that are widely used today.

\subsection{Zero-order Reverse Filtering Algorithm}

Adopting fixed-point iterations for defiltering, our zero-order reverse filter algorithm
is summarized in Alg. \ref{alg:defilter}.

\begin{algorithm}
    \SetKwInOut{Input}{INPUT}\SetKwInOut{Output}{OUTPUT}
    \Input{$\mathbf{J}^*$, $f(\cdot)$, $N^{\textrm{iter}}$}
    \Output{$\mathbf{X}^*$}
    \BlankLine
    $\mathbf{X}^0 \gets \mathbf{J}^*$ \;
    \For {t:= 1 {\bf to} $N^{\textrm{iter}}$ }{
        $\mathbf{X}^{t} \gets \mathbf{X}^{t-1}+(\mathbf{J}^*-f(\mathbf{X}^{t-1}))$
    }
    $\mathbf{X}^* \gets \mathbf{X}^{N^{\textrm{iter}}}$
    \BlankLine
    \caption{Zero-order Reverse Filtering}
    \label{alg:defilter}
\end{algorithm}

\vspace{-0.1in} \paragraph{Implementation Remarks} It is noticeable that the implementation of the method is simple without the need to take derivatives. There are two things to pay attention: the implementation of $f$ and determining $N^{\textrm{iter}}$. Some minor complications of $f$ may have a significant impact on the final result. For instance, applying a Gaussian filter with small spatial support range would lead to significantly worse results, as Eq.~\ref{eqn:gauss} no longer holds for a truncated Gaussian. As we will shown in Sec.~\ref{sec:quantitative}, there is no single optimal iteration number for all filters, thus it needs to be set empirically.

\begin{figure}[t]
\centering
    \includegraphics[width=1.0\linewidth]{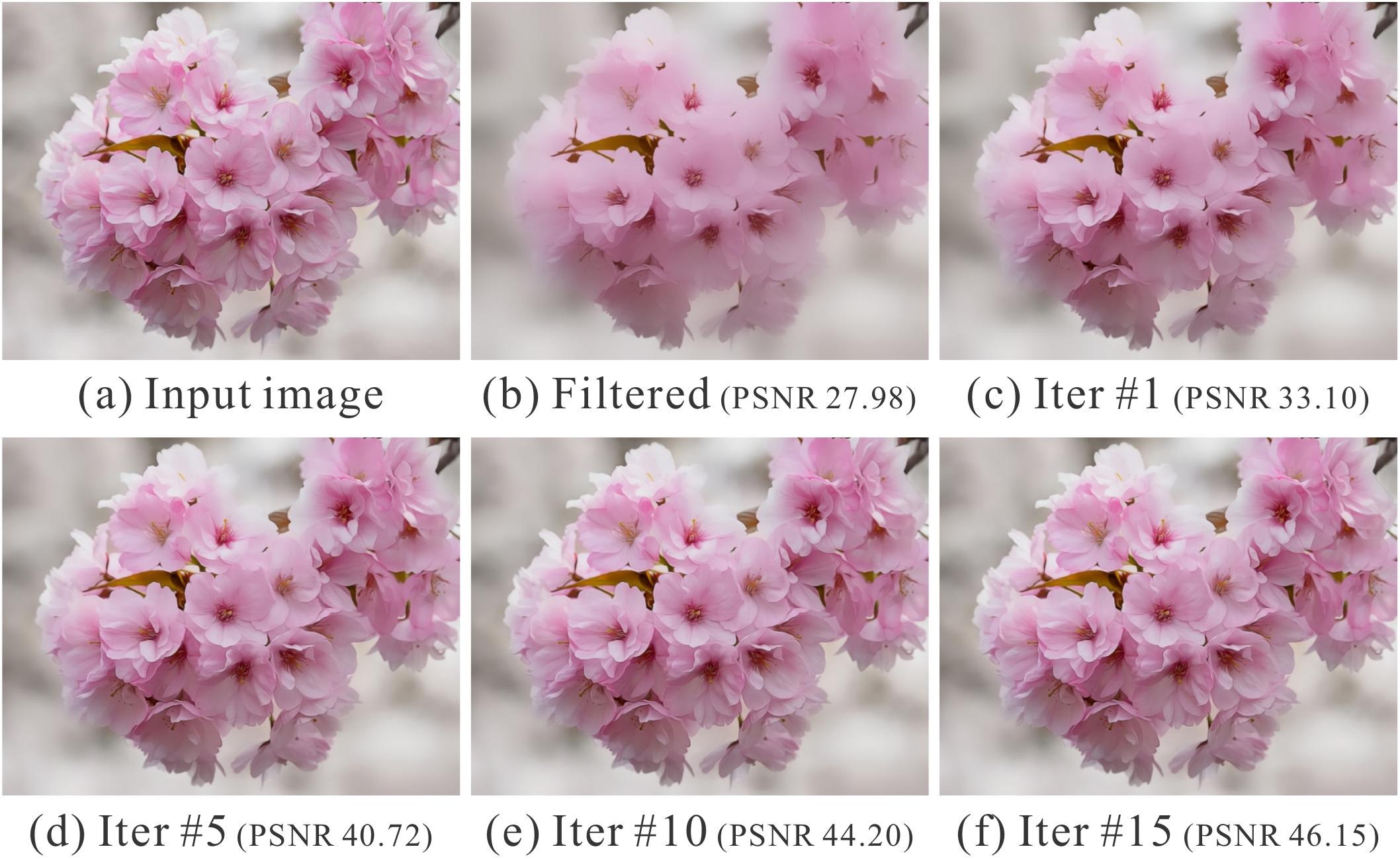}
\caption{Results from iterations of reversing adaptive-manifold filter
\cite{Gastal2012adaptive}.}
    \label{fig:filter_iter}
\end{figure}

In Fig.~\ref{fig:filter_iter}, we show the intermediate results of this iterative process
for reversing adaptive-manifold filter \cite{Gastal2012adaptive}. Applying this filter
reduces PSNR to 27.98dB compared to the original image. Applying our DeFilter with 15
iterations raises the PSNR to 46.15dB -- there is almost no noticeable difference between
the original and defiltered images. Many more examples are provided in the paper and in
the supplemental material to manifest its generality.

Our proposed method also has other advantages besides simplicity. It is parameter free.
The only parameter is the number of iterations, which can be fixed in prior. It also
works well on many nonlinear and complicated filters, such as bilateral filter
\cite{Tomasi1998bilateral} and guided filter \cite{He2010guided}. Even
global-optimization-based methods \cite{Farbman2008wls} can be reversed effectively.

\vspace{-0.1in} \paragraph{Differences from Other Residual Strategies} Residual-based
enhancement was used for other tasks before -- unsharp mask enhances images by adding
edge residuals. In super-resolution, the well-known back-projection technique
\cite{irani1991improving} iteratively refines high-res images by back projecting errors.
Our method is fundamentally different from these strategies. Unsharp mask is a
single-step process and does not recover true details. On the other hand, back-projection
is more like a gradient descent method based on the imaging model. Our method does not
need to know exact filter model, as illustrated in Fig.~\ref{fig:app_retouch} and
following examples.

\begin{table*}
    \center
    \caption{Evaluation Results on BSD 300 dataset (PSNR)}\label{tab:evaluate}
    \vspace{0.1in}
    \scriptsize
    \begin{tabular}{ c | c | c | c | c |  c | c | c | c | c | c | c | c | c | c | c | c }
    \hline
    Filter      & GS  & BF  & BFG & BFPL& GF  & AMF & RF  & TF  & RGF & MF  & WMF & BM3D& L0  & RTV & WLS & LE     \\  \hline\hline
    Init GT     &27.75     &25.50     &25.67     &27.85     &28.21     &27.36     &27.80     &28.75     &30.06     &26.01     &33.00
                &32.27     &26.86     &25.34     &24.80     &29.30   \\  \hline
    Final GT    &\bf{41.70}&\bf{45.28}&35.78     &32.97     &\bf{51.05}&47.82     &28.63     &27.61     &\bf{44.20}&N/A       &22.98
                &\bf{38.84}&28.32     &\bf{30.27}&28.64     &7.96    \\  \hline
    Best GT     &\bf{41.70}&\bf{45.28}&\bf{37.70}&\bf{36.92}&\bf{51.05}&\bf{48.10}&\bf{31.16}&\bf{29.21}&\bf{44.20}&\bf{26.07}&\bf{35.43}
                &\bf{38.84}&\bf{28.99}&\bf{30.27}&\bf{29.71}&\bf{44.21}   \\  \hline\hline
    Init DT     &36.62     &33.12     &32.72     &33.36     &32.49     &30.66     &29.11     &26.85     &37.27     &37.08     &35.72
                &38.61     &31.70     &34.84     &29.08&37.27   \\  \hline
    Final DT    &\bf{79.07}&\bf{80.50}&64.21     &54.32     &\bf{87.94}&53.02     &\bf{60.22}&36.00     &\bf{45.67}&N/A       &35.65
                &\bf{57.98}&36.51     &\bf{45.38}     &62.20&45.55   \\  \hline
    Best DT     &\bf{79.07}&\bf{80.50}&\bf{66.39}&\bf{54.82}&\bf{87.94}&\bf{53.26}&\bf{60.22}&\bf{36.28}&\bf{45.67}&\bf{38.49}&\bf{39.16}
                &\bf{57.98}&\bf{38.04}&\bf{45.38}&\bf{62.28}&\bf{45.67}   \\  \hline
    \end{tabular}\\
    \vspace{0.05in}
    \scriptsize {\textbf{GS}: Gaussian Filter,
                 \textbf{BF}: Bilateral Filter \cite{Tomasi1998bilateral},
                 \textbf{BFG}: Bilateral Grid \cite{chen2007real},
                 \textbf{BFPL}: Permutohedral Lattice \cite{adams2010fast},
                 \textbf{GF}: Guided Filter \cite{He2010guided},
                 \textbf{AMF}: Adaptive Manifold Filter \cite{Gastal2012adaptive},
                 \textbf{RF}: Domain Transform \cite{Gastal2011domain},
                 \textbf{TF}: Tree Filter \cite{tip-14-linchao-bao},
                 \textbf{RGF}: Rolling Guidance Filter with AMF \cite{Zhang2014rgf},
                 \textbf{MF}: Median Filter,
                 \textbf{WMF}: Weighted Median Filter \cite{Zhang2014wmf},
                 \textbf{BM3D}: BM3D Denoise \cite{dabov2007bm3d},
                 \textbf{L0}: L0 Smooth \cite{Xu2011l0},
                 \textbf{RTV}: Relative Total Variation \cite{Xu2012ts},
                 \textbf{WLS}: Weighted Least Square \cite{Farbman2008wls},
                 \textbf{LE}: Local Extrema \cite{subr2009edge}.}
\end{table*}

\section{Experiments}\label{sec:exp}
We conduct many experiments to evaluate the effectiveness and the generality of our
method on various common filters. Our experiments are conducted on a PC with an Intel
Xeon E5 3.7GHz CPU. We use the authors' implementations for all filters. In order to
demonstrate the restoration ability, parameters of these filters are purposely set to
large values to produce strong filtering effect. Both qualitative and quantitative
results are provided for comparison.

\subsection{Quantitative Evaluation}\label{sec:quantitative}
 To quantitatively evaluate the restoration accuracy,
we build a dataset of 300 images with large appearance and structural variation based on
BSD300 from Berkeley segmentation dataset \cite{MartinFTM01}. We apply our method to
reverse 16 famous filters, which we believe cover the majority of practical ones. We fix
the number of iterations to 50, relatively large for the purpose of analyzing
convergence. Our method is initialized as $\mathbf{J}^*$.

Two similarity measures based on PSNR are used. {\it Ground truth} ($\textbf{GT}$) error
measures the difference between the recovered image and the unfiltered original image.
{\it Data-term} ($\textbf{DT}$) error calculates the difference between input filtered
image and refiltered version of the recovery result. We calculate initial PSNRs before
the first iteration (\textbf{Init}), final PSNRs after 50th iteration (\textbf{Final}).
Considering that complicated filters may violate the ``reversible condition'', we also
report the best PSNRs (\textbf{Best}) achieved in the entire process. The results are
listed in Table \ref{tab:evaluate}.

\begin{figure}[t]
    \centering
        \includegraphics[width=1.0\linewidth]{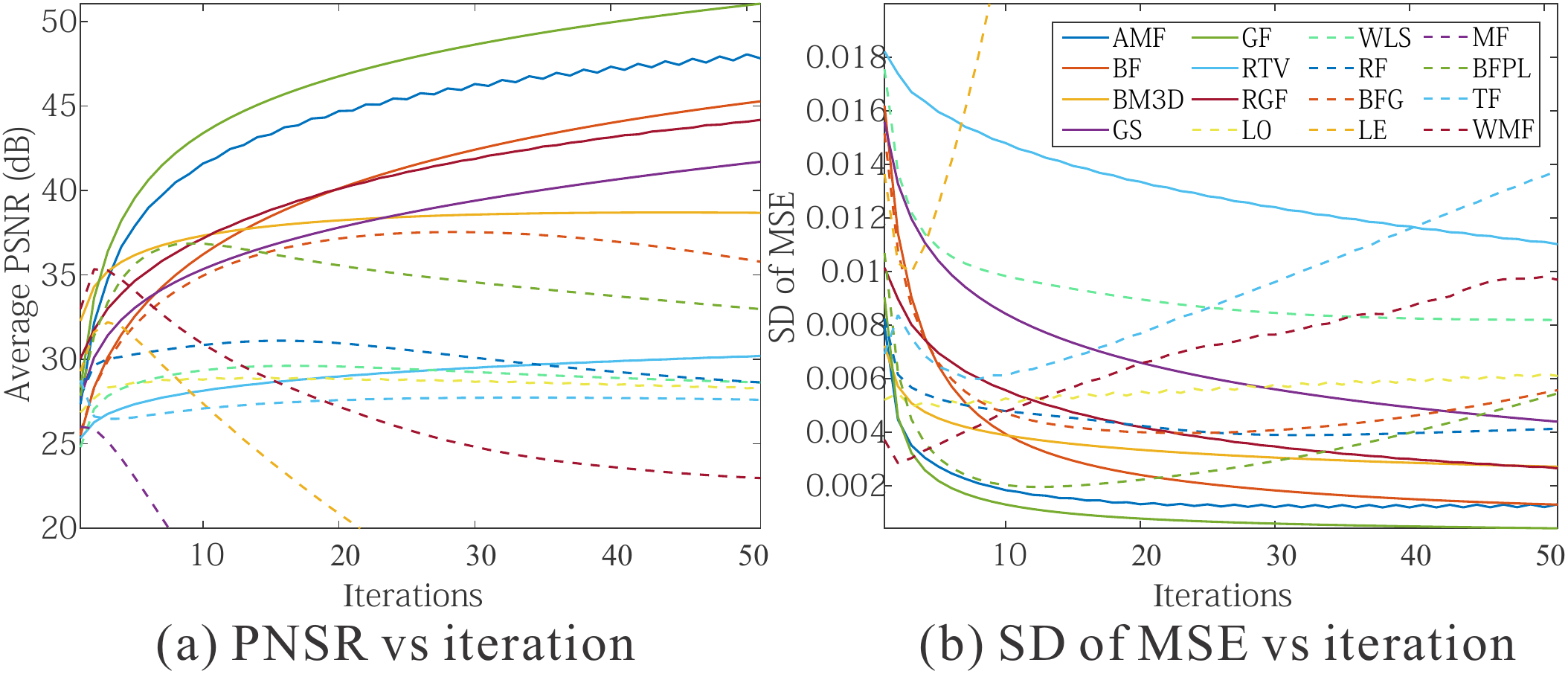}
    \caption{Curves of (a) PSRN vs. iteration and (b) Standard deviation (SD) of mean
            square error (MSE) vs. iteration for common filters.}
    \label{fig:exp_psnr_iter}
\end{figure}

\subsection{Result Analysis} We make several important observations from the results in
Table \ref{tab:evaluate}. First, $\textbf{DT}$ PSNRs are generally larger than
$\textbf{GT}$ ones, which complies with the fact that our method is basically a feedback
system based on $\textbf{DT}$ errors. Second, a larger $\textbf{DT}$ PSNR does not
necessarily correspond to a larger $\textbf{GT}$ PSNR. For lossy filters, such as median
filter (MF) and local extrema filter (LE), the same output can be obtained from different
inputs. Thus the defiltered image may not be the same as the original one, which makes
perfect sense.

To analyze the convergence of our method on different filters, we plot the
PSNR-vs-iteration curves and the curves of standard deviation (SD) of mean square error
(MSE) vs. iteration in Fig.~\ref{fig:exp_psnr_iter}. For filters that are well
reversible, including Gaussian filter (GS), bilateral filter (BF), guided filter (GF),
adaptive manifold filter (AMF), rolling guidance filter (RGF), BM3D and relative total
variation (RTV), the PSNRs consistently increase.

\begin{figure}
\centering
    \includegraphics[width=1.0\linewidth]{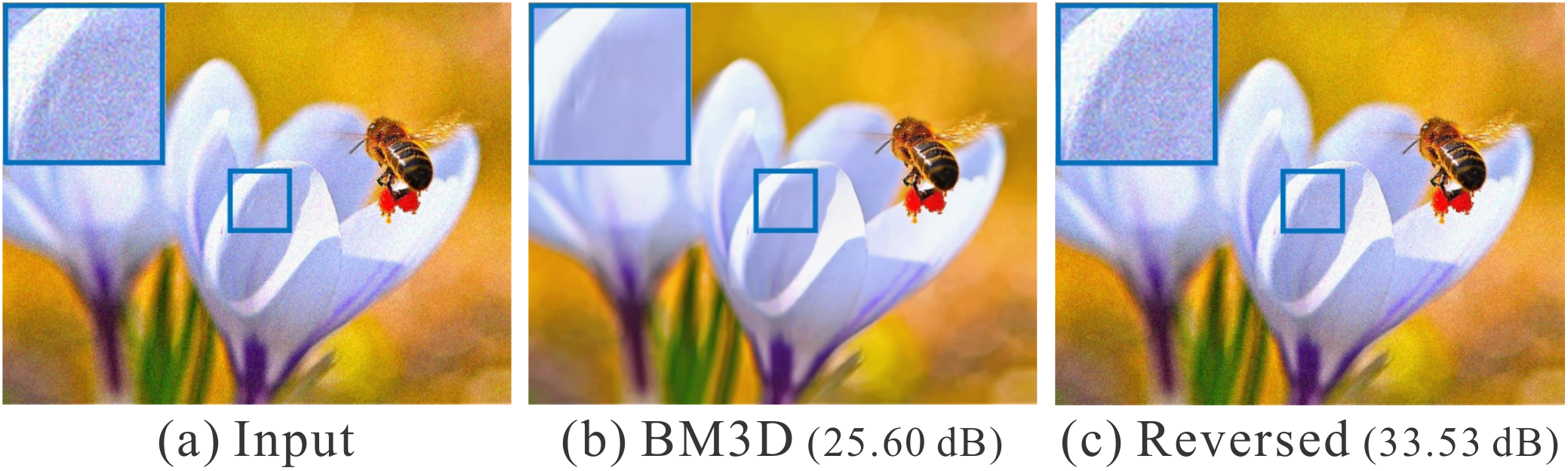}
\caption{{\bf Image detail recovery.} (a) Original noisy input image. (b) Filtered image by
BM3D. (c) Defiltered image using our method. Note that the noise patterns between input
and our recovered images are very close.}\label{fig:exp_noise}
\end{figure}

For filters that are partially reversible, such as bilateral grid (BFG), permutohedral
lattice (BFPL), domain transform (RF), tree filter (TF), L0 smooth (L0) and weighted
least square (WLS), PSNRs increase in early iterations, and then decrease or oscillate in
later ones. This complies with our previous theoretical analysis that reversible
components dominate images. Thus a good number of, by default 10, iterations can yield
satisfying results for most filters.

Finally, for filters that are discontinuous in many places, such as median filter (MF),
weighted median filter (WMF), and local extrema filter (LE), our method does not work
very well with slightly increased PSNRs in the first a few iterations. For these filters,
we claim them as {\it not reversible} by our algorithm.

\begin{figure}[t]
    \centering
        \includegraphics[width=1.0\linewidth]{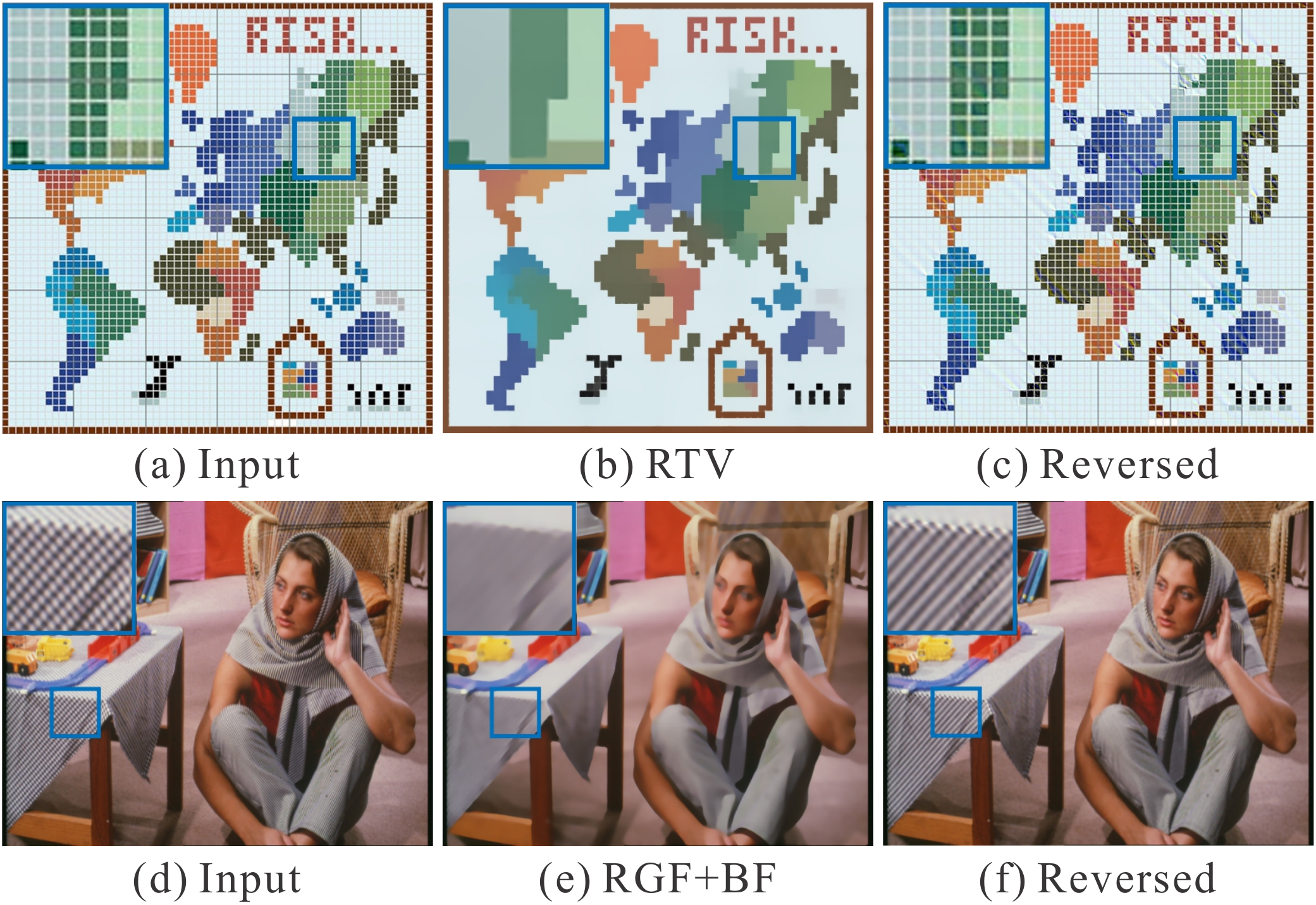}
\caption{Our method can recover even small-scale image texture by reversing
detail-removal filters.}\label{fig:app_texture_scale}
\end{figure}

\subsection{Recovered Detail Verification}

Are the details recovered by our method true signals? In Fig.~\ref{fig:exp_noise}, we
apply BM3D denoising on an image, and then use our method to reverse it. Visual
inspection indicates that the noise patterns of the defiltered image in
Fig.~\ref{fig:exp_noise}(c) is very similar to those of the original image in
Fig.~\ref{fig:exp_noise}(a).

In Fig.~\ref{fig:app_texture_scale}, we first apply RTV and RGF+BG filters to remove
small scale texture and a level of image structures from the input images, then use our
method to recover them. Again, our method takes back the small-scale details, consistent
with the input. This is because most image filters largely suppress, but do not
completely remove these details. The signal residual, albeit not visually prominent, can
still help recover the original input. Naturally, if some image structures are completely
destroyed in filtering, they cannot be recovered well. This explains the difference of
the tablecloth patterns in Fig.~\ref{fig:app_texture_scale}(f) and (d).

\vspace{0.1in}
\subsection{Robustness}

\begin{figure}[t]
    \centering
        \includegraphics[width=1.0\linewidth]{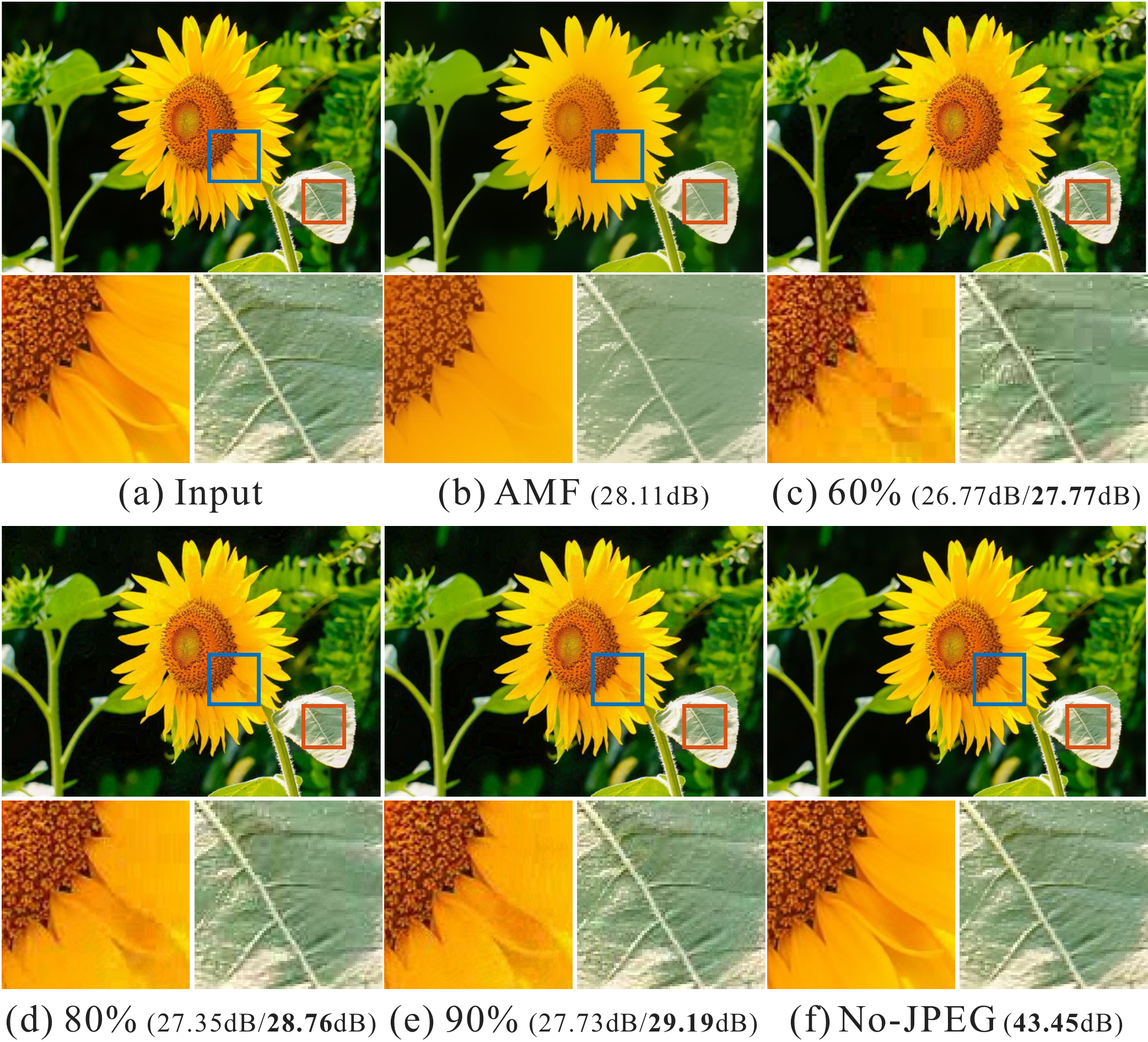}
    \caption{{\bf Reverse filtering under JPEG compression.} (a) Original image.
        (b) Filtered image using AMF.
        (c)-(e) Our recovered images in 20 iterations under different levels of
        JPEG compression. Compressed input PSNR ({\bf left}) and recovered result PSNR ({\bf right}).
        (f) Our DeFilter result of (b).
        }\label{fig:exp_jpeg}
\end{figure}

We also evaluate the robustness of our method against lossy JPEG compression applied to
the filtered image. In Fig.~\ref{fig:exp_jpeg}, we first apply AMF to suppress weak edges
and texture, followed by standard JPEG compression (MATLAB JPEG encoder with quality
60\%, 80\%, 90\%). The DeFilter results shown in Fig.~\ref{fig:exp_jpeg}(c)-(e) contain
sufficiently recovered image details even under lossy compression.

\vspace{0.1in}
\subsection{More Results and Applications}\label{sec:results} 

\begin{figure*}[th]
    \centering
        \includegraphics[width=1.0\linewidth]{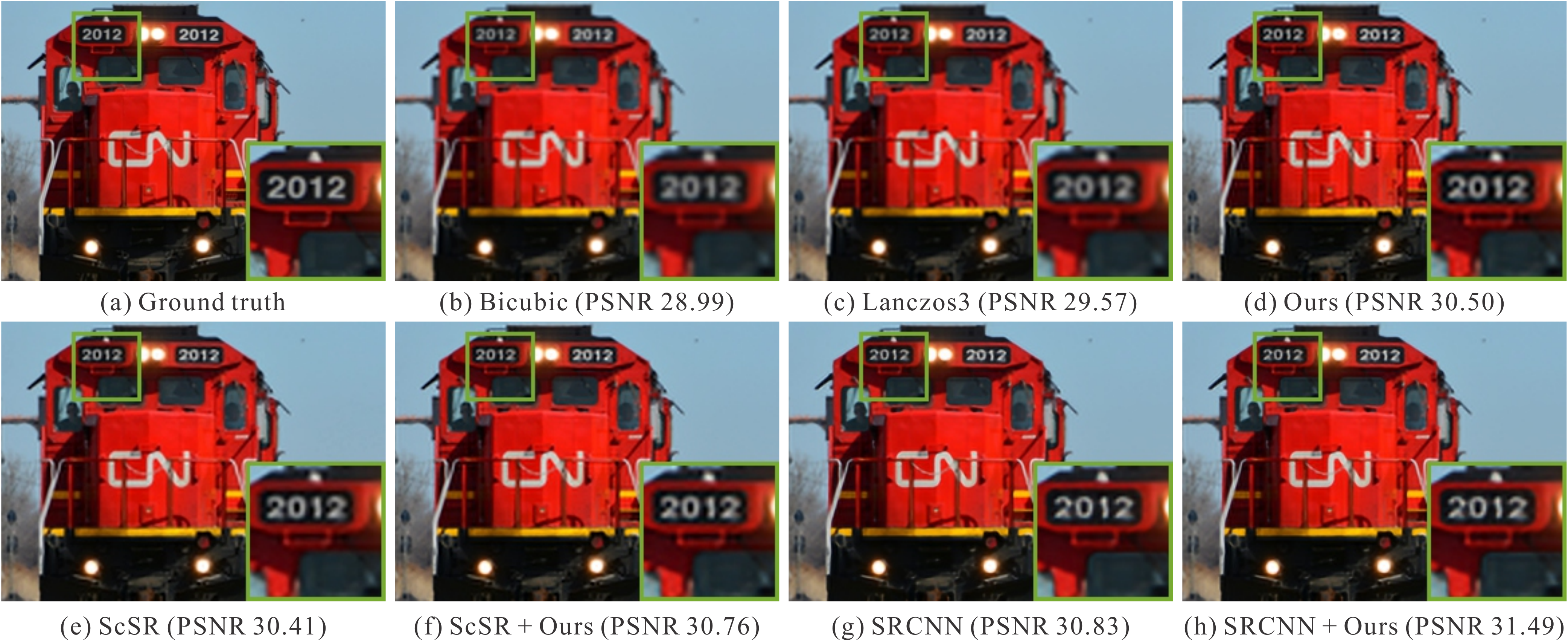}
    \caption{{\bf Zero-order super-resolution.} (a) Ground truth. (b)-(c)
        Bicubic and Lanczos3 interpolation. (e)\&(g) Results of ScSR \cite{yang2008image}
        and SRCNN \cite{dong2014learning}. (d)\&(f)\&(h) Ours results with (b)\&(e)\&(g) as
        initialization, respectively. }\label{fig:app_sr_compare}
\end{figure*}

\vspace{-0.1in}
\paragraph{Zero-order Super-resolution}
Single image super-resolution is ill-posed due to information loss. Interestingly, if we
define the low-res image generation process as a special down-sampling filter with scale
factor $\sigma=2$:
\begin{align}
    f_{SR}(\mathbf{I})=\text{resize} ( \text{resize} ( \mathbf{I}, 1/\sigma ), \sigma ),
\end{align}
we can apply our reverse filtering to rebuild the high-res image. One example is shown in
Fig.~\ref{fig:app_sr_compare}. Bicubic and Lanczos3 interpolation ($\times2$) achieves
PSNRs of 28.99dB and 29.57dB, respectively. Directly applying our method using the
Bicubic result as initialization achieves 30.50dB, which is on par with more
sophisticated learning-based methods, such as ScSR \cite{yang2008image} and SRCNN
\cite{dong2014learning}. Note that our method does not rely on external data or parameter
tweaking. Moreover, if we use the results of ScSR \cite{yang2008image} and SRCNN
\cite{dong2014learning} as our initialization, we can further improve them as shown in
Fig.~\ref{fig:app_sr_compare}(f)(h). Therefore, our method can be used as {\it a general
low-cost post-processing} for improving existing super-resolution methods. We would like to point out that the amount of improvement our method can achieve is image-structure-dependent, thus varies with different input images.

\vspace{-0.1in}
\paragraph{Zero-order Nonblind Deconvolution}

\begin{figure}[t]
    \centering
        \includegraphics[width=1.0\linewidth]{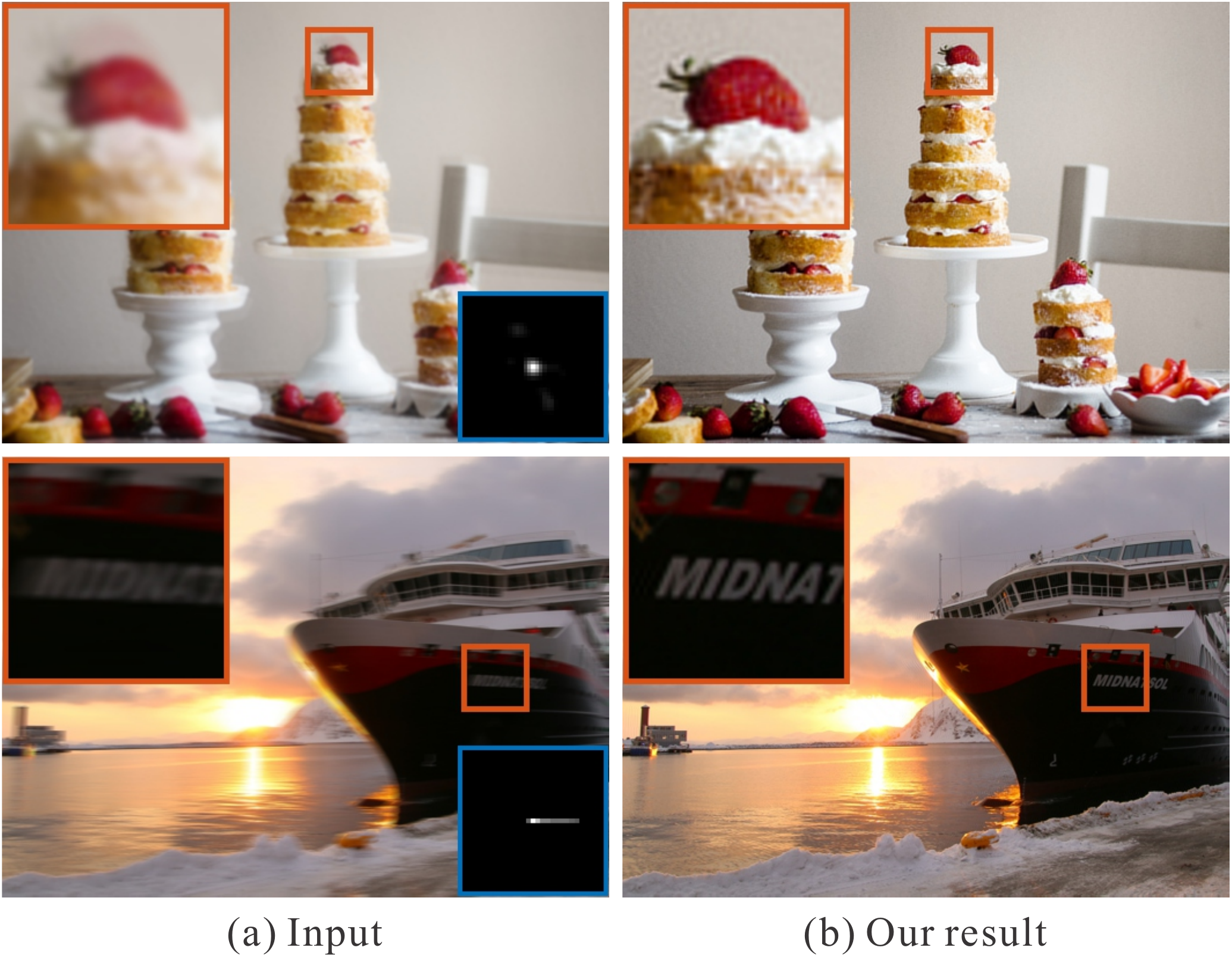}
    \caption{{\bf Zero-order nonblind deconvolution.} (a) Input image with known blur kernel.
        (b) Recovered sharp image using our method.}\label{fig:app_deconv}
\end{figure}

Image convolution is in the common linear form $ f_{Conv}(\mathbf{I}) =
\mathbf{I}\otimes\mathbf{K}$ that our method can handle. In Fig.~\ref{fig:app_deconv}, we
apply our method to image deconvolution. Different from other non-blind deconvolution
methods, the blur kernel may not be known as long as blur effect can be re-applied. Our
method works well for kernels with few zero/negative components in frequency domain to avoid
severe information loss and to satisfy Eq.~(\ref{eq:freq_cond}).

\vspace{-0.1in}
\paragraph{Applications Against Visual Deceiver}

Post-process for hand-held device Apps to beautify human faces and improve image quality
is ubiquitous. These filters hide a lot of information that a picture originally capture,
which can be sometimes regarded as ``visual deceiver''. Our method is the first in its
kind to reverse this post-process and show the original look without needing to know the
filter that the Apps implemented. Two examples are shown in Fig. \ref{fig:app_retouch}
and more in our supplementary material due to page limit.

\begin{figure}[t]
\centering
    \includegraphics[width=1.0\linewidth]{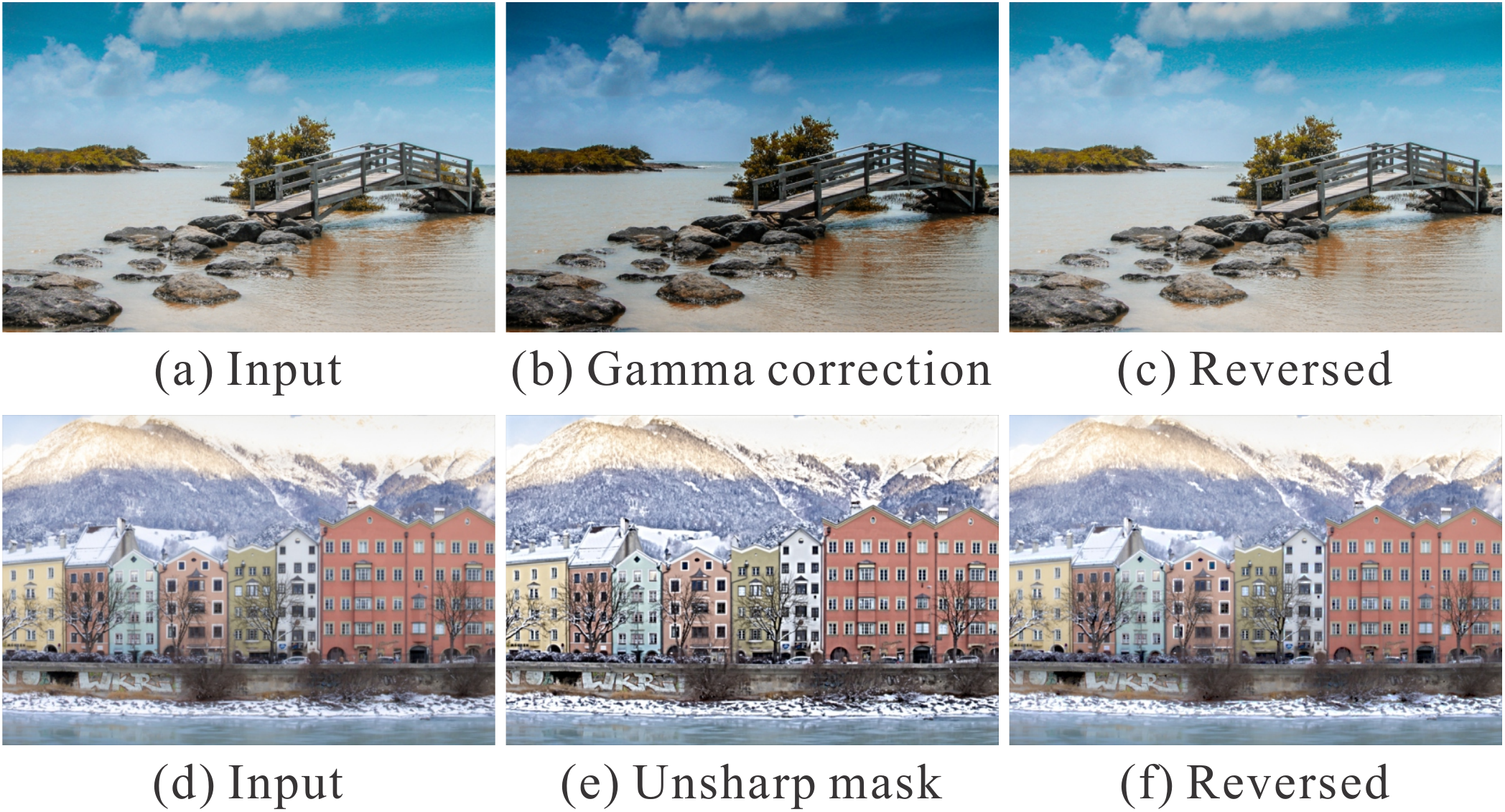}
    \caption{Zero-order restoration for gamma correction and unsharp mask.}\label{fig:gamma_unsharp}
\end{figure}

\vspace{-0.15in}
\paragraph{Reversal of Other Operators} Previous discussions mostly
focus on image smoothing processes. Our method can also work well on some
different operators. Fig.~\ref{fig:gamma_unsharp}(a)-(c) show an example
of gamma correction reverse. Since gamma correctin is basically an element-wise
monotonous operation (invertible), it can be easily verified using our iterative method.
Fig.~\ref{fig:gamma_unsharp}(d)-(f) show the reversed effect of unsharp mask
sharpening method. With mild sharpening parameters, this process also keeps
main structures and energies unchanged, which ensures the correctness of our method.

\vspace{-0.1in}
\section{Concluding Remarks}
\vspace{-0.1in}
We have tackled an unconventional problem of reversing general filter. We have analyzed the
condition that a filter can be reversed, and proposed a zero-order reverse filtering
method based on fixed-point iterations. Extensive experiments show that this simple method can work well for a wide variety of common filters.

While our method has demonstrated its notable simplicity and generality, it has several
limitations. Firstly, depending on filter complexity, the effectiveness may vary. For
instance, reversing the median filter is not easy, as reported in
Table~\ref{tab:evaluate}. Secondly, our method cannot bring back details that are
completely lost during filtering, as shown in Figs.~\ref{fig:app_texture_scale}(f) and
\ref{fig:exp_jpeg}(c).

{\small
\bibliographystyle{ieee}
\bibliography{frev}
}

\end{document}